# GEOMIR2K9 - A Similar Scene Finder

Alwin de Rooij
Leiden University

**Abstract**

The main goal of the GEOMIR2K9 project is to create a software program that is able to find similar scenic images clustered by geographical location and sorted by similarity based only on their visual content. The user should be able to input a query image, based on this given query image the program should find relevant visual content and present this to the user in a meaningful way. Technically the goal for the GEOMIR2K9 project is twofold. The first of these two goals is to create a basic low level visual information retrieval system. This includes feature extraction, post processing of the feature data and classification/ clustering based on similarity with a strong focus on scenic images. The second goal of this project is to provide the user with a novel and suitable interface and visualization method so that the user may interact with the retrieved images in a natural and meaningful way.

## 1. Introduction

Our goal is to find similar scenic images clustered by geographical location and sorted by similarity based only on their visual content. The user should be able to input a query image, based on this given query image the program should find relevant visual content and present this to the user in a meaningful way. Technically the goal for the GEOMIR2K9 project is twofold. The first of these two goals is to create a basic low level visual information retrieval system. This includes feature extraction, post processing of the feature data and classification/ clustering based on similarity with a strong focus on scenic images. The second goal of this project is to provide the user with a novel and suitable interface and visualization method so that the user may interact with the retrieved images in a natural and meaningful way. There has been significant prior research [1-13] which has explored a wide variety of approaches from Markov random walks [1] to fusion of features [13].

## 2. Method and Ideas

To extract low level features that can give meaningful information for a similarity measure aimed at retrieving scenic (and outside) images taken at a multitude of locations from all over the world the following steps were taken. The images are first resized to a standard size to give representative counts when generating the histograms (explained below). Pictures that were made on the side were detected and adjusted accordingly resulting in pictures of size 640x480 and 480x640 pixels. All these images were taken from the MIR2009 database.

One of the ideas implemented in the GEOMIR2K9 project is to have one simple low level feature extraction method that contains related information about the image. Often different feature extraction methods are used that independently extract information about the color, texture and shape of an image. To bridge this I came up with an edge direction color histogram, a feature extraction method where Sobel edge detection is used to extract

8 areas of directions an edge can have and a class of non edges. These different edge features all contain a quantized color histogram relating color and texture (and in a way shape) directly to each other. The color space of these images is in RGB. To gain perceptually better results for the edge direction color histogram the RGB values are converted to the XYZ color space and from there to the perceptual uniform CIE L*A*B* color space. The conversion from XYZ to CIE L*A*B* is done with the parameters observer = 2° and illuminant = D65. D65 is created to represent lighting conditions outside all over the world, which fits this project very well considering the contents of the MIR2009 database. Quantization takes place after converting the colors to CIE L*A*B*. The L* (lightness) is not used in the histograms.

The features extracted from the database go through a post processing stage. The first part is to remove redundant features from the feature set to prevent unnecessary calculations at a later stage by simply threshold it per feature. So if a certain feature for all images has a value below the threshold the feature is removed. This is done with the idea in mind that the used database is representative for the subject of experimentation of this project. The second post processing stage is of a slightly different nature. When analyzing the features it was obvious that some colors were there in very large quantities, but were not necessarily more important in perception than some other less obvious features. To prevent that these would mask other important feature details the following post processing method was devised.

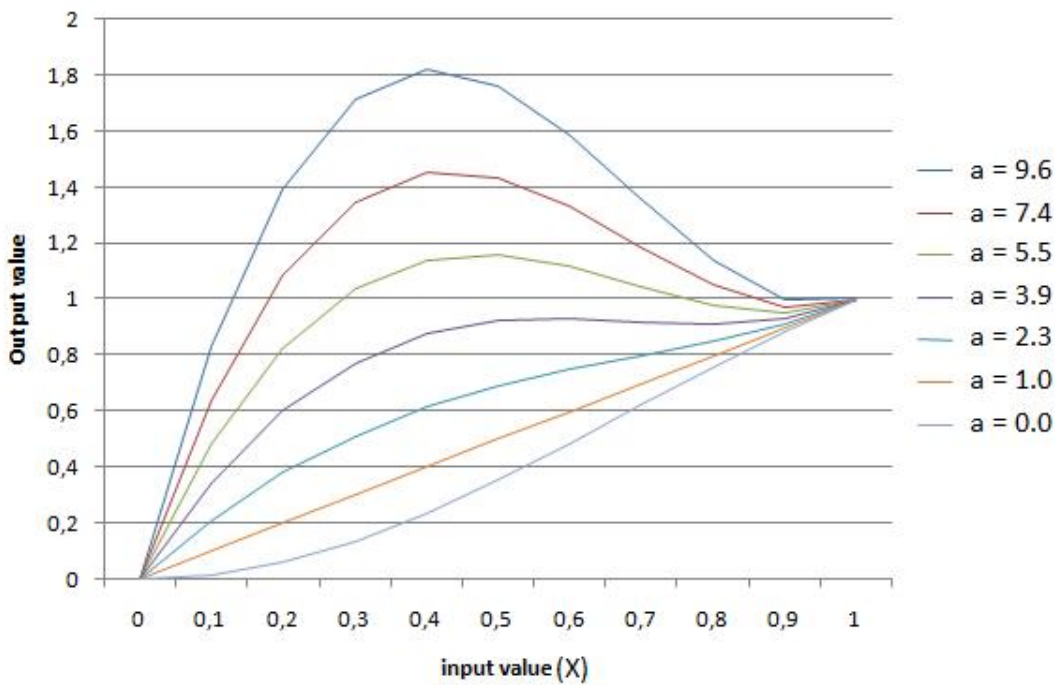

Fig 1: a graph demonstrating the effect of 'a' on incoming data 'X'

The features are first normalized. These normalized features are then processed using:

$$X_{new} = (a-1)*(1-sin(X*PI/2))+1$$

Where 'a' is the factor that determines the exaggeration of the middle features and X is the normalized feature. After experimentation a = 2.3 was used. Using this process it is possible to make less obvious features more important.

After the feature extraction and post processing the resulting features are classified using a Kohonen Neural Network (Self Organizing Map). Before training the neural network the feature data is normalized again. In the training phase of the neural network weight vectors slowly converge into vectors representing groups of images containing certain prominent features. After training all images in the MIR2009 database are assigned to the weight vector in the neural network they are most similar to, generating clusters of similar images. These are later used to gene clusters similar to the features of a query image. The similarity measure used is the Euclidean distance.

As stated above the second main goal of this research project is to create an interface that enables the user to query the database with an image and retrieve the data clustered on geographical location and sorted by similarity. The interface itself is kept very basic. It includes only a button to open and query an image, a button to locate the needed feature files and a button to locate the MIR2009 database. More important is the method used to visualize and present the images retrieved from the database.

To achieve this a physics based approach is taken for the visualization of visually similar image cluster. More specific the approach is a force directed graph layout inspired by the Kawada-Kawai and Fruchterman Rheingold methods. When a set of similar clusters of images is retrieved based on a query image the program goes to a set of steps. The first step is to determine the different countries that belong to the images. Simulations of particles are then generated, one for every country. These particles are then connected to the middle of the screen with a simulation of a spring. After this all particles representing a country are set to repulse the other country particles. After this first step it is determined whether there are and if so what cities belong to these countries from the image cluster. Based on this particles are generated, connected with strings to a country representing particle. Also all city particles are set to repulse each other. The last step is to set particles for the images related to the cities and countries. This is done in the same fashion as the other steps, but now thumbnails of the images are loaded and visualized on the screen. On simulation the force directed graph self organizes in such a way that all images are clustered based on their geographical location. To sort the images on similarity images belonging to clusters within a certain radius to the most similar cluster are drawn first on the screen, images from the most similar clusters are drawn last letting the most similar images appear on top.

## 3. Discussion and Future Work

This approach to sorting images on similarity is somewhat general in the sense that the program does not know the most similar image but only the group of most similar images. Further interaction with the visualization is possible. On mouse over on an image the image blows up and its geographical location is shown bottom left. There is also the possibility to drag images away if a cluster of images is to dense. The visualization will directly self organize.

## Appendix

Using GEOMIR2K9:
1) Make sure the program knows where to find the relevant data:
    a) Select the directory containing the MIR2009 database by clicking:

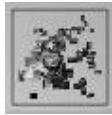

b) (Only needed when the program asks for it) Select the directory containing the data (som) files. This is in our case "./data/". Alternative usable data can be found in the "other/alternative data/" folder.

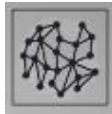

c) If all are set correctly you can query an image

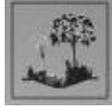

2) If an image is queried correctly (this will take 10 to 90 seconds), the visualization appears.
3) If your mouse is above a picture it will blow up and its geographical location will show bottom left. Dragging pictures to reorganize the visualization a bit is also possible.
4) if something is or went wrong you will get an error message explaining what's up.

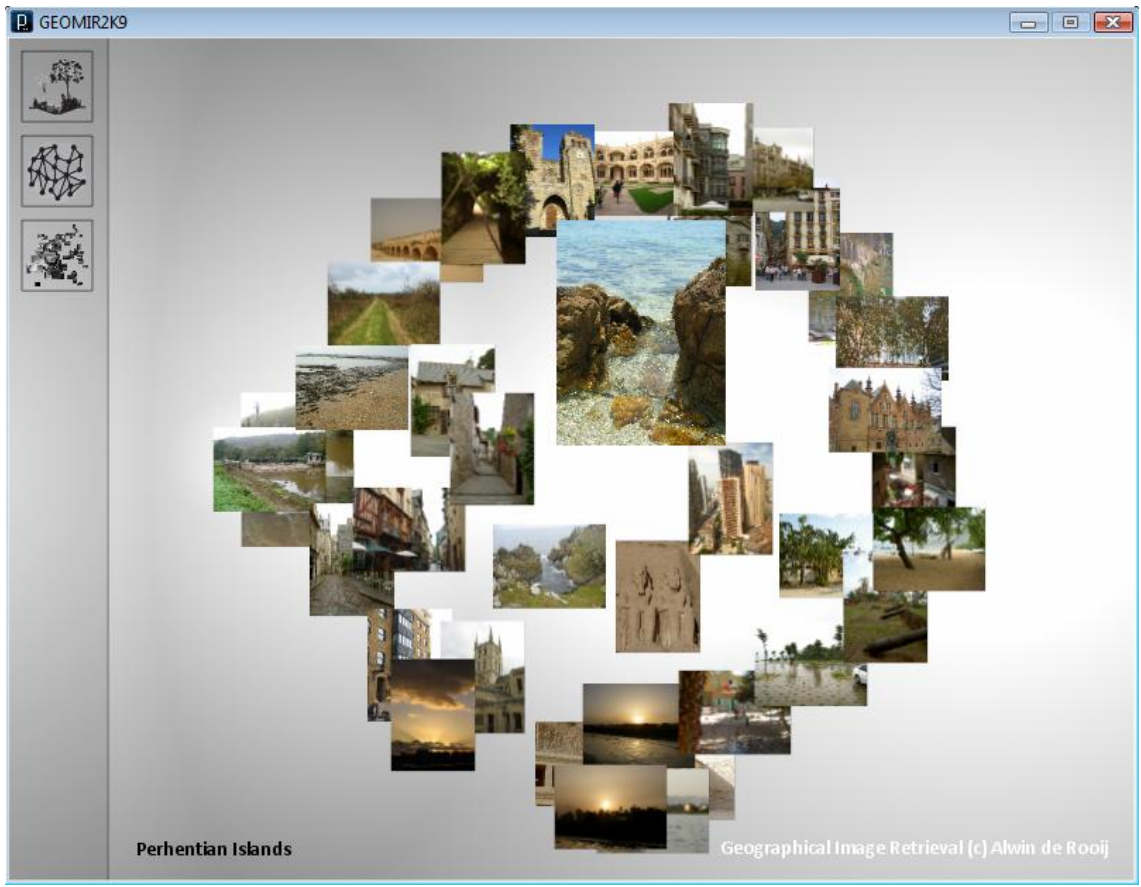

Using other/FEATURE_EXTRACTION:

➔ This program is used in the processing.org program to generate a file (FEATURES.txt) containing the edge direction color histogram values. The files is written to the "./data/" folder. The MIR2009 database must be set by clicking on the screen. This program is not user friendly and not intended to be so!

Using other/CLUSTERING:
➔ This program is used in the processing.org program to generate the files SOM.txt, CLASSIFICATION.txt and STRUCTURE.txt in its data folder. FEATURES.txt generated by the FEATURES program needs to be present in the "./data/" folder beforehand. The generated files are used by GEOMIR2K9. The MIR2009 database must be set by clicking on the screen. This program is not user friendly and not intended to be so!